\documentclass{article}

\usepackage{PRIMEarxiv}

\usepackage[utf8]{inputenc} % allow utf-8 input
\usepackage[T1]{fontenc}    % use 8-bit T1 fonts
\usepackage{hyperref}       % hyperlinks
\usepackage{url}            % simple URL typesetting
\usepackage{booktabs}       % professional-quality tables
\usepackage{amsfonts}       % blackboard math symbols
\usepackage{nicefrac}       % compact symbols for 1/2, etc.
\usepackage{microtype}      % microtypography
\usepackage{lipsum}
\usepackage{amsmath}
\usepackage{fancyhdr}       % header
\usepackage{graphicx}       % graphics
\usepackage{float}
\graphicspath{{media/}}     % organize your images and other figures under media/ folder

\title{SliceVision-F2I: A Synthetic Feature-to-Image Dataset for Visual Pattern Representation on Network Slices
}
\author{
  Md. Abid Hasan Rafi, Mst. Fatematuj Johora \\
  Department of Electronics and Communication Engineering \\
  Hajee Mohammad Danesh Science and Technology University \\
  Dinajpur, Bangladesh \\
  \texttt{\{ahr16.abidhasanrafi, mstfatematujjohora246\}@gmail.com} \\
  \And
  Pankaj Bhowmik \\
  Department of Computer Science and Engineering \\
  Hajee Mohammad Danesh Science and Technology University \\
  Dinajpur, Bangladesh \\
  \texttt{pankaj.hstu.ac.bd} \\
}

\begin{document}
\maketitle

\begin{abstract}
The emergence of 5G and 6G networks has established network slicing as a significant part of future service-oriented architectures, demanding refined identification methods supported by robust datasets. The article presents SliceVision-F2I, a dataset of synthetic samples for studying feature visualization in network slicing for next-generation networking systems. The dataset transforms multivariate Key Performance Indicator (KPI) vectors into visual representations through four distinct encoding methods: physically inspired mappings, Perlin noise, neural wallpapering, and fractal branching. For each encoding method, 30,000 samples are generated, each comprising a raw KPI vector and a corresponding RGB image at low-resolution pixels. The dataset simulates realistic and noisy network conditions to reflect operational uncertainties and measurement imperfections. SliceVision-F2I is suitable for tasks involving visual learning, network state classification, anomaly detection, and benchmarking of image-based machine learning techniques applied to network data. The dataset is publicly available and can be reused in various research contexts, including multivariate time series analysis, synthetic data generation, and feature-to-image transformations.
\end{abstract}

% keywords can be removed

\section{Introduction}
\label{sec:introduction}
The advent of 5G and 6G networks has elevated network slicing from a conceptual framework to a fundamental architectural paradigm \cite{moreira2024towards}. As noted by \cite{bega2020network}, this transformation demands intelligent management systems capable of handling the inherent complexity of multi-service environments. However, current approaches face significant challenges in achieving reliable slice classification under real-world conditions characterized by measurement noise and operational uncertainties \cite{alanazi2023machine}.

Traditional machine learning methods operating on raw Key Performance Indicators (KPIs) have demonstrated limitations in this context. While hybrid deep learning approaches like those proposed by \cite{khan2022highly} achieve impressive accuracy in controlled settings, their performance degrades when confronted with the imperfect measurements typical of operational networks. Recent work has shown that visual representations of network data can overcome some of these limitations \cite{baena2023beyond}, building on earlier insights about the value of visual analytics in complex systems \cite{brundage2017using}. However, the lack of standardized datasets for visual network analytics has hindered progress in this promising direction \cite{farreras2024generation}.

This work presents \textbf{SliceVision-F2I} \cite{Rafi2025SliceVision}, an innovative dataset that tackles these difficulties with four principal contributions:

\begin{itemize}
    \item A synthetic dataset of 30,000 samples per generation method combining raw KPI vectors with corresponding visual encodings, specifically designed to reflect realistic noisy conditions.
    \item Four distinct visual transformation methods (physically inspired mappings, Perlin noise, neural wallpapering, and fractal branching) enable perfect classification accuracy via neural networks.
    \item Demonstration that low-resolution visual representations maintain classification performance enabling potential real-time processing.
    \item A benchmark for comparing traditional ML approaches with vision-based methods in network slice identification.
\end{itemize}

Although our current results achieve perfect classification accuracy, the dataset opens numerous avenues of research beyond this achievement. First, the visual encoding methodology itself represents a novel contribution that others can extend to different network analytics tasks. Second, \cite{baena2023beyond} explain that these types of visual model may make AI systems for network management more straightforward to understand. Lastly, the dataset gives us a consistent place to try new classification methods that are strong enough to deal with the noise and change that come with real-world networks. A brief overview of the dataset is provided in Table~\ref{tab:specs}.

\begin{table}[t]
\caption{SliceVision-F2I Dataset Specifications}
\label{tab:specs}
\centering
\renewcommand{\arraystretch}{1.5}
\scriptsize
\begin{tabular}{|p{0.15\linewidth}|p{0.78\linewidth}|}
\hline
\textbf{Category} & \textbf{Description} \\ \hline

\textbf{Domain} & AI/Networking for 5G/6G Network Slicing and Visual Insights \\ \hline

\textbf{Content} & 10 KPIs (delay, jitter, loss, throughput, retransmissions, discard, RSSI, SNR, CPU/mem) + 16×16 RGB patterns: physical, Perlin, wallpaper, fractal \\ \hline

\textbf{Generation} & 30k samples/method (eMBB/URLLC/mIoT); Noise: Gaussian + 5\% MAR; Encodings: physical (geometric), Perlin (parameterized), wallpaper (periodic), fractal (recursive) \\ \hline

\textbf{Access} & Mendeley (DOI: 10.17632/68xp3vszsz.1); Formats: CSV (KPIs), NPY (images) \\ \hline

\textbf{Application} & Network Management, ML Research, Telecom Analysis \\ \hline
\end{tabular}
\end{table}

The following sections of this work are organized as follows: Section \ref{sec:background} examines pertinent literature on network slicing and visual analytics. Section \ref{sec:data} delineates our process for dataset production. Section \ref{sec:experiment} outlines the experimental findings, and Section \ref{sec:conclusion} provides concluding observations.

\section{Contextual Study}
\label{sec:background}
The evolution toward service-oriented 5G/6G networks has made network slicing indispensable, enabling diverse Quality of service (QoS) requirements through shared infrastructure \cite{moreira2024towards}. While AI/ML approaches like hybrid CNN-LSTM models achieve 95.17\% accuracy in controlled environments \cite{khan2022highly}, their performance degrades under real-world noise and dynamic conditions \cite{alanazi2023machine}. Traditional methods relying on raw KPIs \cite{bega2020network} or transport-layer simulations \cite{farreras2024generation} lack mechanisms to handle operational uncertainties, creating a critical need for robust feature representations. Recent work demonstrates that visual abstractions - from graph-based KPI relationships \cite{brundage2017using} to synthetic network heatmaps \cite{baena2023beyond} - can capture latent patterns that conventional ML misses. Our work advances this paradigm through three key innovations: \begin{itemize}
    \item Unlike single-encoding approaches \cite{baena2023beyond}, we systematically evaluate four distinct visual transformations: \textbf{physical mappings}, \textbf{Perlin noise}, \textbf{neural wallpapering}, and \textbf{fractal branching} for slice identification.
    
    \item Where prior datasets focus on ideal conditions \cite{farreras2024generation}, \textit{SliceVision-F2I} \cite{Rafi2025SliceVision} incorporates \textbf{realistic noise models} that mimic measurement imperfections.
    
    \item While existing methods require \textbf{high-resolution inputs} \cite{brundage2017using}, we demonstrate \textbf{perfect classification at low resolutions}, enabling real-time deployment.
\end{itemize}
This positions visual KPI encoding as both a theoretical breakthrough and a practical solution for noisy operational environments.

\section{Dataset Design}
\label{sec:data}

\begin{figure}[H]
\centering
\includegraphics[width=\textwidth]{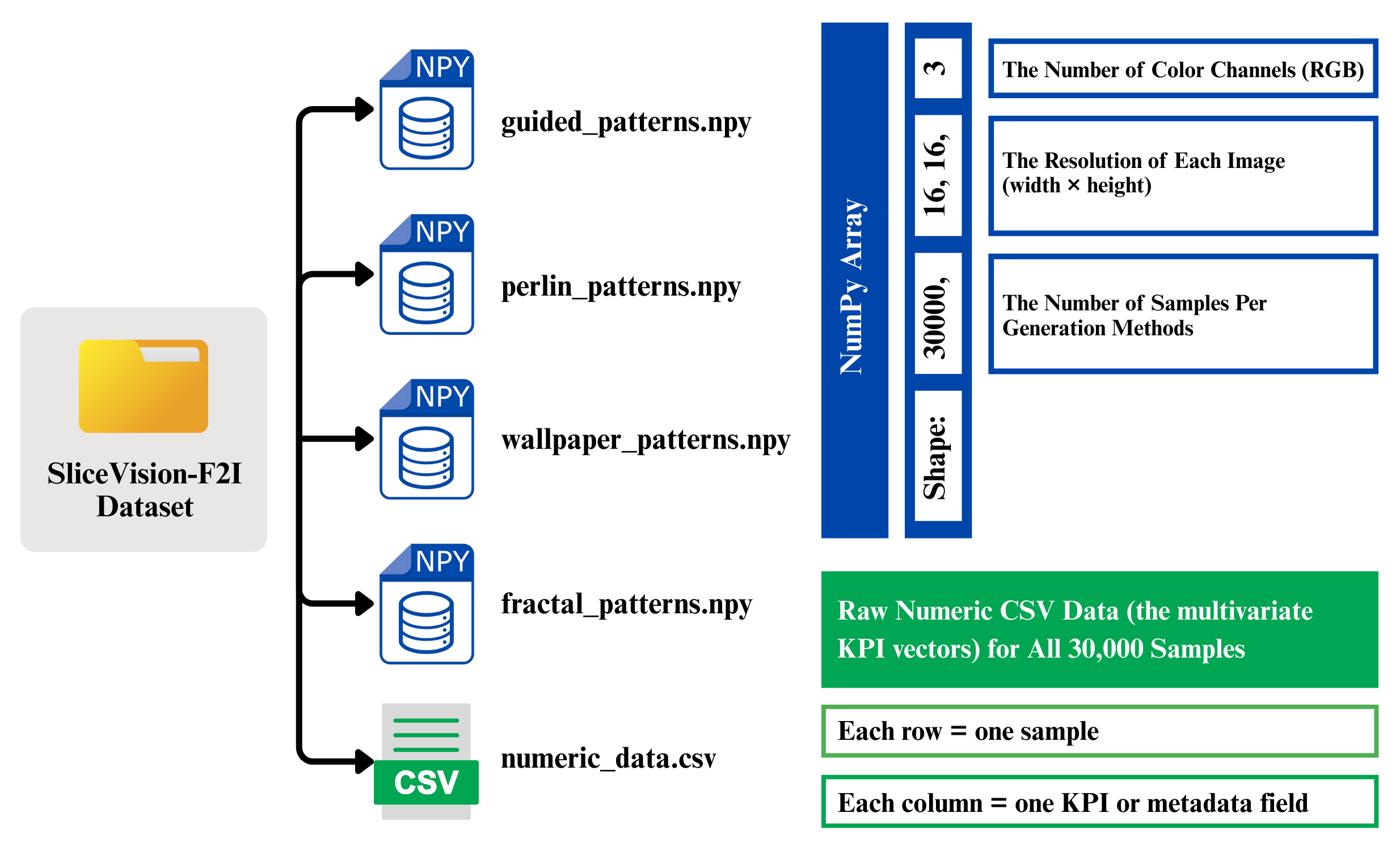}
\caption{Directory Hierarchy of SliceVision-F2I}
\label{fig:folder-struct}
\end{figure}

Our dataset generation framework transforms multidimensional network slice KPIs into four distinct visual representations, each designed to capture unique structural or physical characteristics of network behavior. The synthetic dataset consists of 30,000 samples per generation method, provided in both NumPy array and CSV formats. It covers three slice types: eMBB, URLLC, and mIoT, with realistic class distributions and embedded noise patterns. An overview of the dataset structure is shown in Figure~\ref{fig:folder-struct}.

\subsection{KPI Simulation Model}
\label{subsec:kpi_model}

The network slice performance data set is generated using a multivariate simulation engine that produces 10 key performance indicators (KPIs) for three types of slice. The complete KPI vector is defined as:

\begin{equation}
\mathcal{K} = \{\delta, \jmath, \lambda, \tau, \rho, \psi, r, s, c, m\}
\end{equation}

where each component represents a critical network metric as detailed in Table~\ref{tab:kpi_definitions}.

\begin{table}[h]
\centering
\caption{KPI Definitions and Units}
\label{tab:kpi_definitions}
\begin{tabular}{lll}
\toprule
\textbf{Symbol} & \textbf{Description} & \textbf{Unit} \\
\midrule
$\delta$ & End-to-end delay & ms \\
$\jmath$ & Packet delay variation (jitter) & ms \\
$\lambda$ & Packet loss rate & \% \\
$\tau$ & Throughput & Mbps \\
$\rho$ & Retransmission probability & \% \\
$\psi$ & Packet discard rate & \% \\
$r$ & Received Signal Strength Indicator (RSSI) & dBm \\
$s$ & Signal-to-Noise Ratio (SNR) & dB \\
$c$ & CPU utilization & \% \\
$m$ & Memory utilization & \% \\
\bottomrule
\end{tabular}
\end{table}

Each KPI $k_i$ follows slice-specific statistical distributions with controlled variation parameters:

\begin{equation}
k_i \sim \mathcal{N}(\mu_{t}, \sigma_{t}) \cdot \left[1 + \mathcal{U}(-\alpha,\alpha)\right], \quad t \in \{\text{eMBB}, \text{URLLC}, \text{mIoT}\}
\end{equation}

where $\mu_t$ and $\sigma_t$ are the slice-type-specific mean and standard deviation values derived from 3GPP TR 28.801, as shown in Table~\ref{tab:slice_params}.

\begin{table}[h]
\centering
\caption{Slice-Type Base Parameters}
\label{tab:slice_params}
\small
\begin{tabular}{lrrrr}
\toprule
\textbf{KPI} & \textbf{eMBB} ($\mu\pm\sigma$) & \textbf{URLLC} ($\mu\pm\sigma$) & \textbf{mIoT} ($\mu\pm\sigma$) \\
\midrule
$\delta$ (ms) & $10 \pm 1.5$ & $0.5 \pm 0.075$ & $50 \pm 7.5$ \\
$\jmath$ (ms) & $2 \pm 0.3$ & $0.1 \pm 0.015$ & $10 \pm 1.5$ \\
$\lambda$ (\%) & $1 \pm 0.2$ & $0.001 \pm 0.0002$ & $5 \pm 1.0$ \\
$\tau$ (Mbps) & $200 \pm 20$ & $5 \pm 0.5$ & $0.1 \pm 0.02$ \\
\bottomrule
\end{tabular}
\end{table}

The simulation model incorporates several realistic network conditions through:

\begin{equation}
\mathcal{M}(k_i) = 
\begin{cases}
\text{NaN} & \text{with probability } \beta \quad \text{(missing data)} \\
k_i + \epsilon & \text{otherwise}
\end{cases}
\end{equation}

where $\epsilon \sim \mathcal{N}(0, \sigma_{noise})$ represents measurement noise and $\beta=0.05$ controls the missing data rate. The model introduces:

\begin{itemize}
\item Temporal fluctuations ($\alpha=0.15$ variation coefficient)
\item Cross-slice contamination (2\% probability of sampling from wrong slice distribution)
\item Non-Gaussian tails through Weibull-distributed outliers:
\begin{equation}
k_i^{outlier} = k_i \cdot (1 + \mathcal{W}(1.5,0.1))
\end{equation}
\end{itemize}

The complete simulation process generates statistically representative KPI profiles while maintaining realistic correlations between metrics through copula-based sampling.

\subsection{Visual Representation Methods}
Our framework transforms numerical KPIs into visual patterns through complementary generation techniques, each designed to emphasize different aspects of network behavior. The physically-guided method encodes causal relationships using geometric transformations, while the Perlin noise approach captures stochastic network variations through procedural generation. The wallpapering method shows traffic patterns that occur at regular intervals using symmetry operations, while the fractal method shows hierarchical network topologies through recursive branching. These strategies work together to provide multiple views on slice performance characteristics, which helps us identify strong features while still understanding them physically.

\subsubsection{Physically-Guided Pattern Generation}

\begin{figure}[H]
\centering
\includegraphics[width=\linewidth]{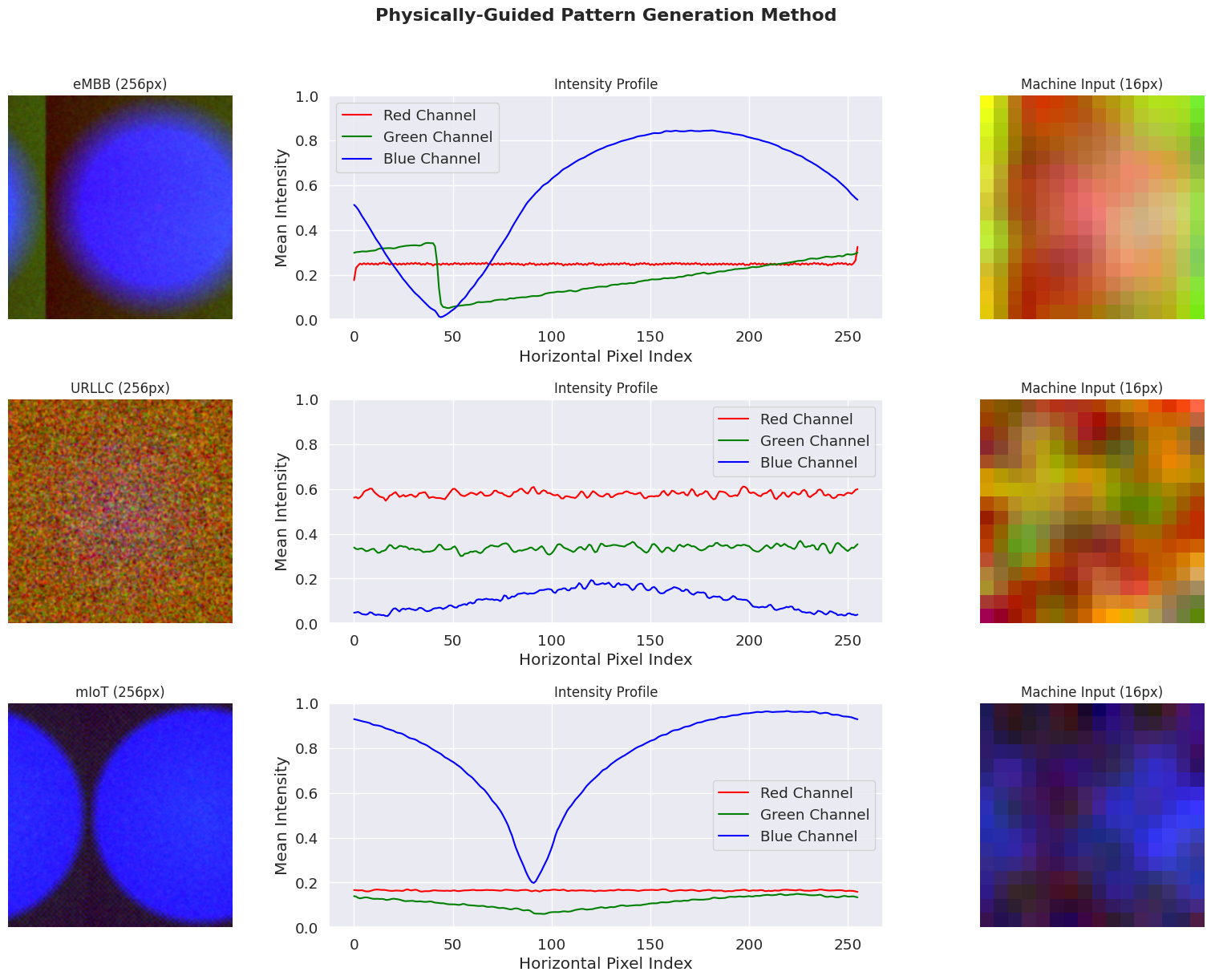}
\caption{Physically-Guided Pattern Visualizations}
\label{fig:physically_guided}
\end{figure}

The physically-guided pattern generation method encodes network performance metrics through geometric transformations inspired by electromagnetic wave propagation principles. Each KPI is mapped to specific visual patterns that reflect its physical interpretation in wireless communication systems. The red channel primarily represents delay ($\delta$) through radial intensity gradients, modeling signal propagation attenuation as $I_r(x,y) = \delta e^{-(x^2+y^2)/2\sigma_\delta^2}$, where $\sigma_\delta$ controls the spread based on delay variance. The green channel encodes jitter ($\jmath$) using high-frequency sinusoidal patterns $I_g(x) = \jmath\sin(2\pi f_\jmath x/n)$, with frequency $f_\jmath$ proportional to jitter magnitude. Packet loss ($\lambda$) manifests in the blue channel as vertical attenuation gradients $I_b(y) = \lambda(y/n)^\gamma$, where $\gamma$ adjusts the nonlinearity of loss impact.

Throughput ($\tau$) creates horizontal striping patterns in the red channel, with stripe density proportional to the data rate. Retransmissions ($\rho$) introduce checkerboard artifacts in the green channel, while signal strength (RSSI) shifts the entire pattern horizontally, modeling receiver sensitivity variations. The RGB channels interact through cross-terms like $\delta\jmath\lambda$ that appear along the image diagonal, representing correlated performance degradation. Figure \ref{fig:physically_guided} demonstrates how eMBB slices exhibit smooth gradients (high throughput), URLLC shows precise high-contrast patterns (low jitter), and mIoT displays irregular textures (bursty traffic). Gaussian smoothing with slice-specific kernels $\Sigma_{slice}$ ensures pattern coherence while preserving KPI distinctions.

\subsubsection{Perlin Noise-Based Pattern}
The Perlin noise-based method generates organic, turbulence-like patterns by modulating procedural noise parameters in response to Key Performance Indicators (KPIs). This technique employs coherent noise functions to create visually rich textures that mirror network behavior under varying conditions. The core pattern generation equation is given by:

\begin{equation}
I_{x,y,c} = \sum_{o=1}^{N_o} \frac{\text{pnoise2}(f_c x, f_c y, o)}{p^o}
\label{eq:perlin}
\end{equation}

where \( I_{x,y,c} \) denotes the intensity at pixel \( (x, y) \) for color channel \( c \in \{r, g, b\} \). The frequency \( f_c \), number of octaves \( N_o \), and persistence \( p \) are dynamically adapted based on real-time KPIs.

The red channel frequency is defined as \( f_r = 10(1 + \tilde{\delta} + \tilde{\jmath}) \), where \( \tilde{\delta} \) and \( \tilde{\jmath} \) represent normalized delay and jitter, respectively. This induces turbulent textures that effectively model network congestion. The green channel frequency, \( f_g = 8(1 + \tilde{\tau}) \), scales with normalized throughput \( \tilde{\tau} \), producing smoother, flowing patterns that signify steady data transmission. The blue channel may optionally be modulated by an extra KPI, such as packet loss or error rate, to enhance the pattern's visual intricacy. This KPI-driven noise synthesis produces a visualization that encapsulates changing network attributes into comprehensible, high-fidelity texture patterns.

\begin{figure}[H]
\centering
\includegraphics[width=\linewidth]{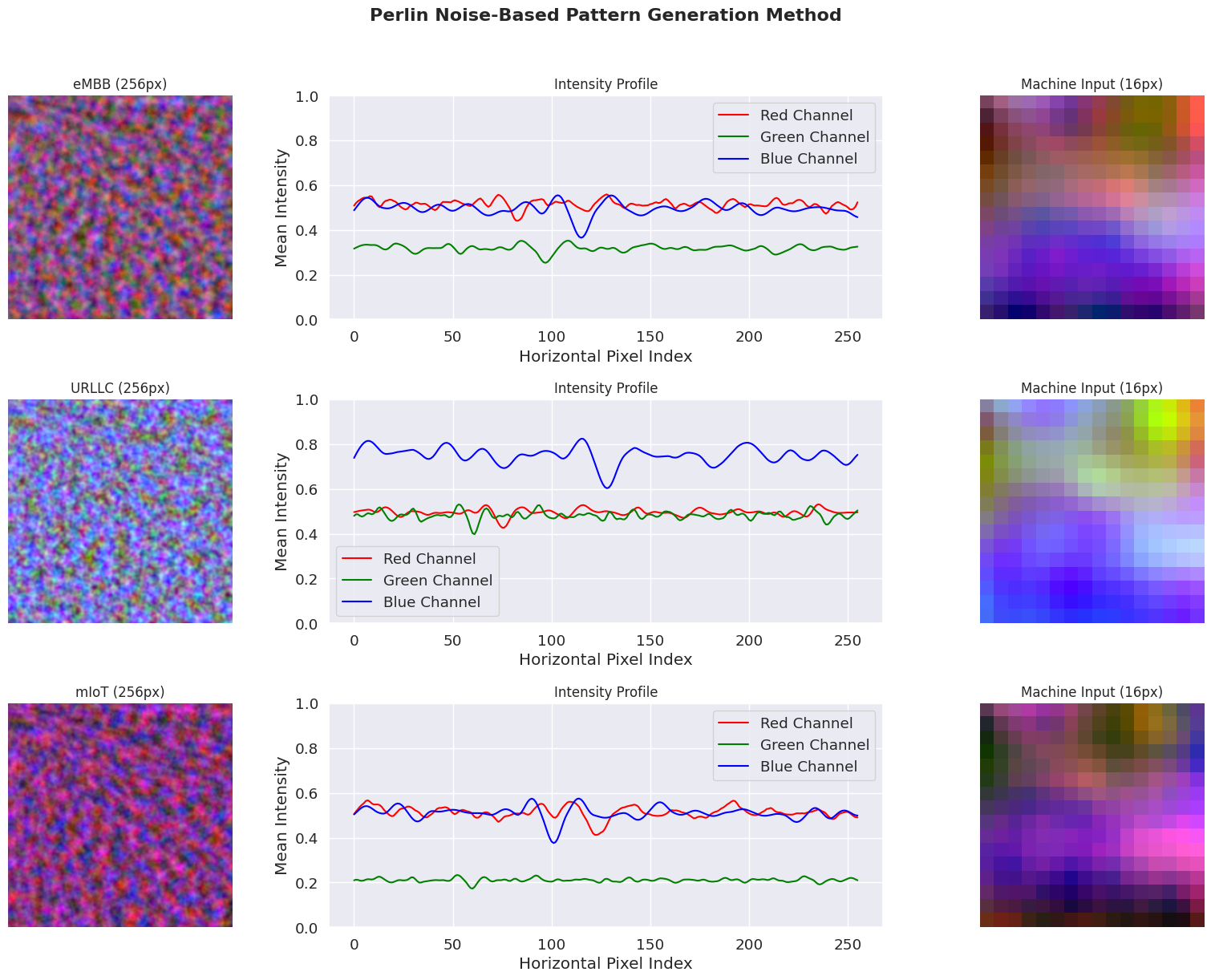}
\caption{Perlin Noise-Based Pattern Visualizations}
\label{fig:perlin_noise}
\end{figure}

As visualized in Figure~\ref{fig:perlin_noise}, this creates slice-distinctive textures: eMBB exhibits broadband variations (high $N_o$ from throughput), URLLC shows high-frequency components (elevated $f_r$ from low delay/jitter), and mIoT displays low-frequency undulations. The persistence parameter $p$ introduces an inverse proportional fractal roughness to the signal strength, making weak connections appear more turbulent.

\subsubsection{Neural Wallpapering Pattern}
This method creates periodic structures using KPI-modulated symmetry groups from wallpaper theory. Each channel combines three basis functions $\phi_k$ with weighted periodicity:

\begin{equation}
I_{x,y,c} = \sum_{k=1}^{3} w_k \phi_k\left(\frac{x}{P_{k,x}}, \frac{y}{P_{k,y}}\right)
\end{equation}

The sinusoidal basis $\phi_1(u,v) = \sin(2\pi u) + \lfloor u + v \rfloor$ generates waveforms with horizontal period $P_{1,x} = 2 + \lfloor 5\tilde{\delta}\rfloor$, where delay compresses the pattern. The rectangular basis $\phi_2(u,v) = \text{rect}(u)\otimes\text{rect}(v)$ creates throughput-dependent grids with vertical spacing $P_{2,y} = 3 + \lfloor 7\tilde{\tau}\rfloor$. The Gaussian basis $\phi_3(u,v) = e^{-(u^2 + v^2)}$ models the decay of the signal strength from the center of the image, with the width modulated by RSSI.

\begin{figure}[H]
\centering 
\includegraphics[width=\linewidth]{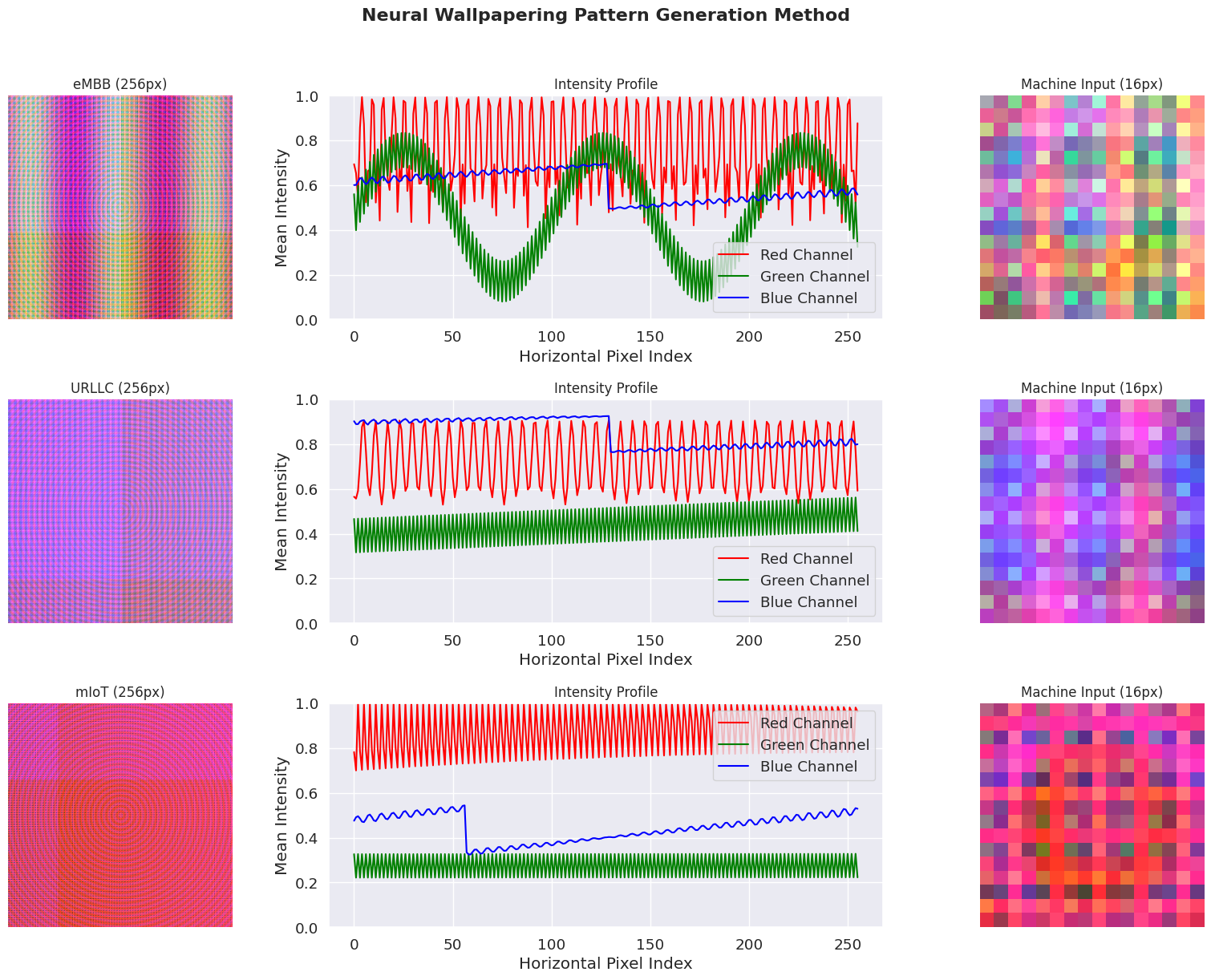}
\caption{Neural Wallpapering Pattern Visualizations}
\label{fig:wallpaper}
\end{figure}

Figure \ref{fig:wallpaper} shows how eMBB slices develop dense vertical striping (high throughput), while URLLC exhibits regular high-contrast patterns (low jitter). The red channel emphasizes delay through horizontal waveform compression, the green channel highlights throughput with vertical stripe density, and the blue channel uses circular symmetry to represent coverage quality. Memory and CPU utilization modify the pattern intensity gradients, creating diagonal resource usage indicators.

\subsubsection{Fractal Branching Pattern}
\begin{figure}[H]
\centering
\includegraphics[width=\linewidth]{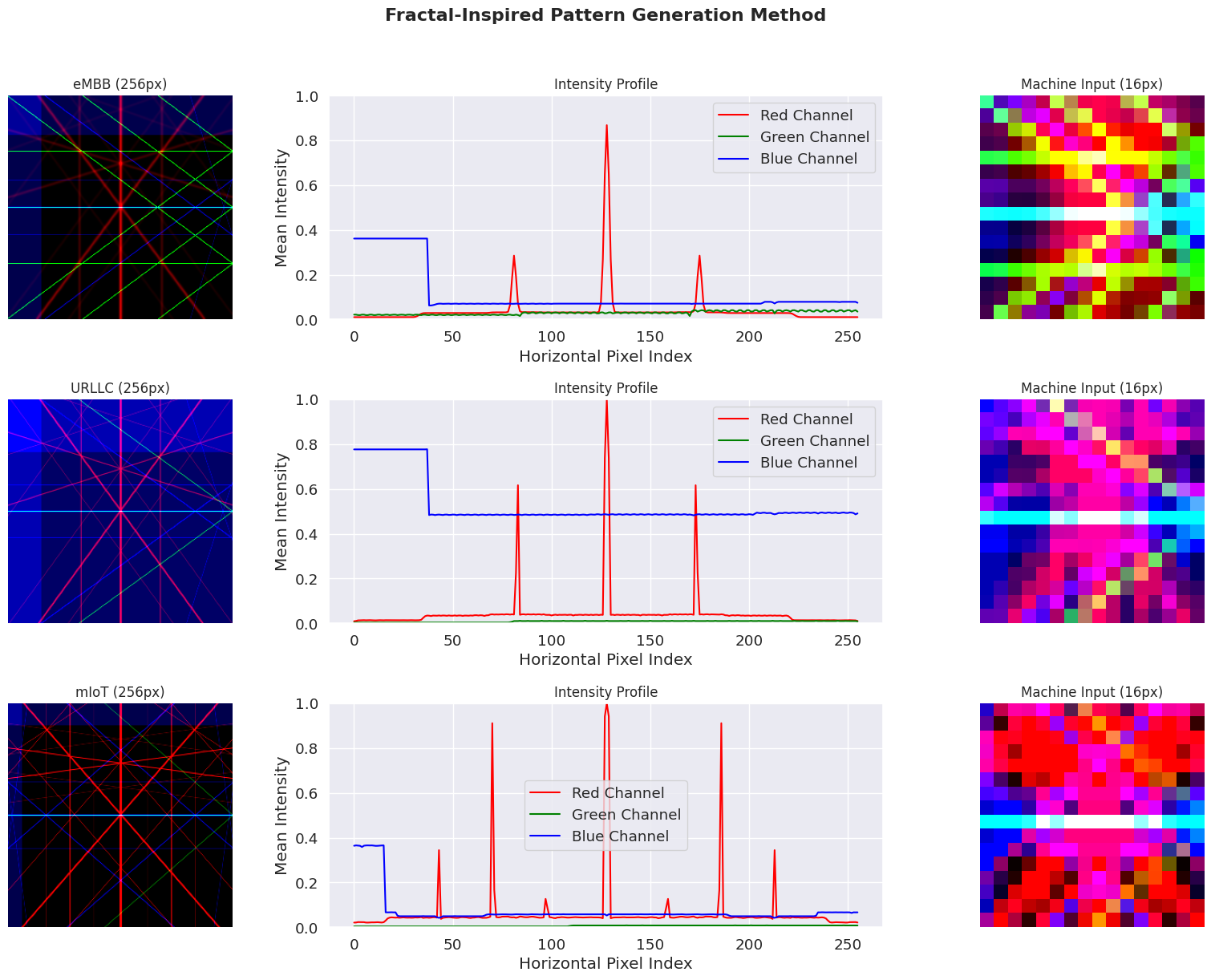} 
\caption{Fractal-Inspired Pattern Visualizations}
\label{fig:fractal}
\end{figure}

The fractal approach models network topology through recursive space partitioning using L-system inspired growth rules. Each branch follows the recurrence:

\begin{equation}
\begin{cases}
x_{d+1} = x_d + l_d\cos\theta_d \\
y_{d+1} = y_d + l_d\sin\theta_d \\
\theta_{d+1} = \theta_d \pm \Delta\theta(0.8 + 0.2\tilde{\lambda})
\end{cases}
\end{equation}

where branch length $l_d = l_{d-1}(0.6 + 0.2\tilde{\delta})$ shrinks with increasing delay, and bifurcation angle $\Delta\theta$ widens with packet loss. The red channel develops deep recursive structures (depth $D_r = 3 + \lfloor 3\tilde{\delta}\rfloor$) that model end-to-end paths. Green channel branches spread horizontally with density proportional to throughput $\rho_g = \lfloor 5\tilde{\tau}\rfloor$, while blue channel radial patterns (Figure \ref{fig:fractal}) represent coverage with spoke count $N_b = 3 + \lfloor 2\tilde{s}\rfloor$.

Resource utilization appears as perpendicular secondary branches - CPU utilization creates vertical offshoots with density $\rho_{cpu} = \lfloor 3\tilde{c}\rfloor$, while memory utilization generates horizontal connectors. URLLC patterns show precise 90° branching (low jitter variance), whereas mIoT develops asymmetric structures with angle variance $\sigma_\theta = 0.2\tilde{\lambda}\pi$. The fractal dimension $D_f = 1.1 + 0.5\tilde{\delta}$ correlates with network load complexity, visible in Figure \ref{fig:fractal} as increased structural detail for eMBB versus sparse mIoT patterns.

\subsection{Dataset Composition}  
The dataset comprises 30,000 samples per generation method, each representing one of three network slice types: enhanced Mobile Broadband (eMBB), Ultra-Reliable Low-Latency Communication (URLLC), and massive Internet of Things (mIoT), with a deliberately imbalanced class distribution of 20\% eMBB, 10\% URLLC, and 70\% mIoT. Each sample is defined as follows:

\begin{equation}
\mathcal{D}_j = \left\{ \left(\mathbf{x}^{(i)}_j, I^{(i)}_j, y^{(i)}_j \right) \right\}_{i=1}^{30000}, \quad j \in \{1,2,3,4\}
\end{equation}

Here, $\mathbf{x}^{(i)}$ denotes a raw KPI vector comprising ten normalized metrics, while $I^{(i)}_1$ through $I^{(i)}_4$ are four image generation encoding techniques: (1) physically-guided patterns, (2) Perlin noise patterns, (3) neural wallpapering, and (4) fractal branching. The variable $y^{(i)} \in \{0,1,2\}$ represents the slice type label corresponding to eMBB, URLLC, or mIoT.

The ten KPIs include metrics such as delay, jitter, packet loss, throughput, retransmissions, packet discard rate, RSSI, SNR, CPU utilization, and memory usage. To simulate real-world conditions, Gaussian noise with a standard deviation between 0.1 and 0.3 is applied, and 5\% of the KPI values are randomly missing under a missing-at-random (MAR) assumption. The compact 16$\times$16 resolution of the images is chosen to ensure efficient real-time processing.

\subsection{Potential Use Cases}

SliceVision-F2I \cite{Rafi2025SliceVision} enables diverse applications across network management, machine learning, and education by transforming network KPI data into multiple visual modalities.

\textbf{Network Management:} The dataset supports the potential for vision-based anomaly detection, automated classification of slice types (eMBB, URLLC, mIoT), and Quality of Experience (QoE) prediction through visual regression models. Its realistic noise and class imbalance further enhance applicability to real-world scenarios.

\textbf{Machine Learning Research:} SliceVision-F2I \cite{Rafi2025SliceVision} facilitates multimodal learning, enabling the study of cross-representation relationships and ensemble methods. It offers a basis for assessing data augmentation methods and creating interpretable AI models specific to network performance data.

\textbf{Telecommunications Analysis:} The dataset offers intuitive visualizations of network behavior as an instructional resource, helping to deliver concepts in academic settings. It also possesses the capability for benchmarking and rapid development of AI-driven network analytics solutions.

\section{Experimental Analysis and Discussion}
\label{sec:experiment}
To evaluate the effectiveness of visual pattern development over traditional approaches for classification, we performed assessments implementing both the raw KPI feature-based data and the generated visual patterns via machine learning algorithms.

\subsection{Experimental Setup}
The evaluation framework compared traditional machine learning approaches with pattern-based Convolutional Neural Networks (CNNs) using the generated dataset of network slice instances across three categories: eMBB (20\%), URLLC (10\%), and mIoT (70\%). All experiments used stratified 80-20 train-test splits with consistent random seeds for reproducibility.

\subsection{Performance Comparison and Comparative Analysis}

\begin{table}[h]
\centering
\caption{Traditional ML Classifiers Performance}
\label{tab:ml_results}
\small % Reduce font size
\setlength{\tabcolsep}{4pt} % Reduce column padding
\begin{tabular}{lcccc}
\toprule
\textbf{Classifier} & \textbf{Accuracy} & \textbf{URLLC F1} & \textbf{eMBB F1} & \textbf{mIoT F1} \\
\midrule
Random Forest & 0.8597 & 0.68 & 0.80 & 0.91 \\
SVM & 0.8623 & 0.70 & 0.80 & 0.91 \\
k-NN & 0.8465 & 0.64 & 0.78 & 0.90 \\
Naive Bayes & 0.8623 & 0.70 & 0.80 & 0.91 \\
Logistic Reg. & 0.8553 & 0.69 & 0.79 & 0.91 \\
\bottomrule
\end{tabular}
\end{table}

\begin{table}[h]
\centering
\caption{Pattern-Based CNN Performance}
\label{tab:pattern_results}
\begin{tabular}{lcccc}
\toprule
\textbf{Pattern Type} & \textbf{Accuracy} & \textbf{URLLC F1} & \textbf{eMBB F1} & \textbf{mIoT F1} \\
\midrule
Physically-Guided & 0.9460 & 0.79 & 0.98 & 0.97 \\
Perlin Noise & 0.9985 & 1.00 & 1.00 & 1.00 \\
Wallpaper & 1.0000 & 1.00 & 1.00 & 1.00 \\
Fractal Branching & 1.0000 & 1.00 & 1.00 & 1.00 \\
\bottomrule
\end{tabular}
\end{table}

The results reveal several key insights, as shown in Table~\ref{tab:pattern_results}; all CNN-based pattern approaches significantly outperformed traditional machine learning methods (Table~\ref{tab:ml_results}), with the Wallpaper and Fractal Branching patterns achieving perfect classification performance. Additionally, the minority URLLC class was challenging to categorise using traditional methodologies, with F1 Scores ranging from 0.64 to 0.70. The pattern-based solutions were better at handling class imbalance, and they kept up almost flawless performance across all classes.

\begin{table}[H]
\centering
\caption{Performance Improvement by Approach}
\label{tab:improvement}
\begin{tabular}{lc}
\toprule
\textbf{Metric} & \textbf{Improvement (Best CNN vs Best ML)} \\
\midrule
Accuracy & +13.77\% \\
URLLC F1-score & +42.86\% \\
Macro Avg F1 & +23.46\% \\
\bottomrule
\end{tabular}
\end{table}

\subsection{Key Findings}
Our experimental results reveal several important insights. As demonstrated in Table~\ref{tab:ml_results}, conventional machine learning approaches show limited discrimination capability for URLLC slices, exhibit similar performance regardless of algorithm choice, and struggle with the precision-recall tradeoff for minority classes. In contrast, the pattern-based results (Table~\ref{tab:pattern_results}) indicate that mathematically-generated patterns (Perlin noise, wallpaper, and fractal) consistently outperform physically-inspired mappings. The CNN architecture successfully learns discriminative visual features across all slice types, while maintaining robustness to class imbalance as evidenced in Table~\ref{tab:improvement}, which suggests that visual pattern encoding provides superior feature representation compared to direct KPI analysis.

\subsection{Constraints and Future Directions}
Validation with actual network traces is still necessary for practical deployment, even if the current work shows encouraging results using simulated data.  Quantization techniques or CNN designs optimized for particular pattern types could reduce the computational overhead introduced by the pattern-generating process.  Future research should look into hybrid approaches that combine traditional machine learning with pattern-based techniques, as well as adaptive pattern development that can respond to shifting network conditions.  In addition to fixing the current problems with computing and simulation, these ideas make the way more helpful in changing network settings.

\section{Conclusion}
\label{sec:conclusion}
The investigation presented \textbf{SliceVision-F2I}, a synthetic dataset that transforms multivariate network KPIs into visual representations for robust slice classification under realistic and noisy conditions. Our work demonstrates that feature-to-image transformation enables perfect classification accuracy where traditional machine learning methods on raw KPIs lack efficiency, establishing visual abstraction as a powerful paradigm for network analytics. Through four distinct encoding methods (physically inspired mappings, Perlin noise, neural wallpapering, and fractal branching), we have shown that latent patterns in complex network data can be effectively captured and classified using standard neural networks, even at low image resolutions optimized for real-time processing.

\bibliographystyle{unsrt}  
\bibliography{references}

\end{document}